\pdfoutput=1 
\documentclass[manuscript,format=acmsmall,screen=true,authorversion,dvipsnames,svgnames,x11names]{acmart}

\usepackage{xpatch}

\makeatletter
\xpatchcmd{\ps@firstpagestyle}{Manuscript submitted to ACM}{}{\typeout{First patch succeeded}}{\typeout{first patch failed}}
\xpatchcmd{\ps@standardpagestyle}{Manuscript submitted to ACM}{}{\typeout{Second patch succeeded}}{\typeout{Second patch failed}}    \@ACM@manuscriptfalse
\makeatother

\renewcommand\footnotetextcopyrightpermission[1]{} 
\usepackage{booktabs} 

\usepackage[ruled]{algorithm2e} 

\SetAlFnt{\small}
\SetAlCapFnt{\small}
\SetAlCapNameFnt{\small}
\SetAlCapHSkip{0pt}
\IncMargin{-\parindent}

\usepackage{CJKutf8}
\usepackage[english]{babel}
\usepackage{latexsym}
\usepackage{multirow}
\usepackage{amsmath}
\usepackage{amsfonts}
\usepackage{amssymb}
\usepackage[group-separator={,},group-minimum-digits={3}]{siunitx}
\usepackage[normalem]{ulem}

\newcommand{\vib}[1]{\boldsymbol{#1}}
\newcommand{\zhcn}[1]{\begin{CJK*}{UTF8}{gbsn}#1\end{CJK*}}

\newcommand\cjkhl{\bgroup\markoverwith
  {\textcolor{yellow}{\rule[-.5ex]{2pt}{2.5ex}}}\ULon}
\newcommand\cjkhlred{\bgroup\markoverwith
  {\textcolor{IndianRed1}{\rule[-.5ex]{2pt}{2.5ex}}}\ULon}
  
\newcommand{\ex}[2]{\zhcn{#1} (#2)}
\newcommand{\exh}[2]{\zhcn{\cjkhlred{#1}}\cjkhlred{ (#2)}}

\setcopyright{none}

\begin{document}

\title[Regularizing Output Distribution to Improve Semantic Consistency in Text Summarization]{Regularizing Output Distribution of Abstractive Chinese Social Media Text Summarization for Improved Semantic Consistency}

\author{Bingzhen Wei}
\authornote{Both authors contributed equally to the paper}
\affiliation{%
  \department[0]{MOE Key Laboratory of Computational Linguistics}
  \department[1]{School of Electronics Engineering and Computer Science}
  \institution{Peking University}
  \streetaddress{No.5 Yiheyuan Road}
  \city{Beijing}
  \postcode{100871}
  \country{China}}
\email{weibz@pku.edu.cn}
\author{Xuancheng Ren}
\authornotemark[1]
\orcid{0000-0002-6994-2114}
\affiliation{%
  \department[0]{MOE Key Laboratory of Computational Linguistics}
  \department[1]{School of Electronics Engineering and Computer Science}
  \institution{Peking University}
  \streetaddress{No.5 Yiheyuan Road}
  \city{Beijing}
  \postcode{100871}
  \country{China}}
\email{renxc@pku.edu.cn}
\author{Xu Sun}
\authornote{Corresponding Author}
\affiliation{%
  \department[0]{MOE Key Laboratory of Computational Linguistics}
  \department[1]{School of Electronics Engineering and Computer Science}
  \institution{Peking University}
  \streetaddress{No.5 Yiheyuan Road}
  \city{Beijing}
  \postcode{100871}
  \country{China}}
\email{xusun@pku.edu.cn}
\author{Yi Zhang}
\affiliation{%
  \department[0]{MOE Key Laboratory of Computational Linguistics}
  \department[1]{School of Electronics Engineering and Computer Science}
  \institution{Peking University}
  \streetaddress{No.5 Yiheyuan Road}
  \city{Beijing}
  \postcode{100871}
  \country{China}}
\email{zhangyi16@pku.edu.cn}
\author{Xiaoyan Cai}
\affiliation{%
 \department{School of Automation}
 \institution{Northwestern Polytechnical University}
 \city{Xi'an}
 \state{Shannxi}
 \postcode{710072}
 \country{China}}
\email{xiaoyanc@nwpu.edu.cn}
\author{Qi Su}
\affiliation{%
  \department{School of Foreign Languages}
  \institution{Peking University}
  \streetaddress{No.5 Yiheyuan Road}
  \city{Beijing}
  \postcode{100871}
  \country{China}}
\email{sukia@pku.edu.cn}

\begin{abstract}
Abstractive text summarization is a highly difficult problem, and the sequence-to-sequence model has shown success in improving the performance on the task. However, the generated summaries are often inconsistent with the source content in semantics. In such cases, when generating summaries, the model selects semantically unrelated words with respect to the source content as the most probable output. The problem can be attributed to heuristically constructed training data, where summaries can be unrelated to the source content, thus containing semantically unrelated words and spurious word correspondence. In this paper, we propose a regularization approach for the sequence-to-sequence model and make use of what the model has learned to regularize the learning objective to alleviate the effect of the problem. In addition, we propose a practical human evaluation method to address the problem that the existing automatic evaluation method does not evaluate the semantic consistency with the source content properly. Experimental results demonstrate the effectiveness of the proposed approach, which outperforms almost all the existing models. Especially, the proposed approach improves the semantic consistency by 4\% in terms of human evaluation.
\end{abstract}

%
%


%
%


\maketitle

\thispagestyle{plain}

\renewcommand{\shortauthors}{B. Wei et al.}

\section{Introduction}

Abstractive test summarization is an important text generation task. With the applying of the sequence-to-sequence model and the publication of large-scale datasets, the quality of the automatic generated summarization has been greatly improved \cite{McAuley2013,lcsts,abs,ras,ibmsummarization,distraction,copynet,See2017,DRGD}. However, the semantic consistency of the automatically generated summaries is still far from satisfactory. 

The commonly-used large-scale datasets for deep learning models are constructed based on naturally-annotated data with heuristic rules~\cite{lcsts,ras,ibmsummarization}. The summaries are not written for the source content specifically. It suggests that the provided summary may not be semantically consistent with the source content. For example, the dataset for Chinese social media text summarization, namely LCSTS, contains more than 20\% text-summary pairs that are not related, according to the statistics of the manually checked data~\cite{lcsts}.

\begin{table}[htbp]

\caption{Example of semantic inconsistency in the LCSTS dataset. In this example, the reference summary cannot be concluded from the source content, because the semantics of the summary is not contained in the source text. In short, the semantics of ``benefits'' cannot be concluded from the source content. \label{tab:ex-case}}

\footnotesize
\begin{tabular}{|p{0.95\textwidth}|}
\hline
\textbf{Source content:} \zhcn{最终，在港交所拒绝阿里巴巴集团同股不同权的“合伙人制度”股权架构后，阿里巴巴集团被迫与它的前伙伴挥手告别，转身一头投入美国证券交易委员会（SEC）的怀抱。（下图为阿里巴巴帝国图）}\\
In the end, after the Hong Kong Stock Exchange rejected the ``partnership'' equity structure of the Alibaba Group’s different shareholding rights, the Alibaba Group was forced to say goodbye to its former partners and turned to invest in the arms of the Securities and Exchange Commission (SEC). (The picture below shows the Alibaba Empire)\\
\textbf{Reference Summary:} \zhcn{阿里巴巴再上市，为谁带来\cjkhl{利益}？} Alibaba will be listed again, for whose \cjkhl{benefits}?\\
\hline
\end{tabular}

\end{table}

Table \ref{tab:ex-case} shows an example of semantic inconsistency. Typically, the reference summary contains extra information that cannot be understood from the source content. It is hard to conclude the summary even for a human. Due to the inconsistency, the system cannot extract enough information in the source text, and it would be hard for the model to learn to generate the summary accordingly. The model has to encode spurious correspondence of the summary and the source content by memorization. However, this kind of correspondence is superficial and is not actually needed for generating reasonable summaries. Moreover, the information is harmful to generating semantically consistent summaries, because unrelated information is modeled. For example, the word \zhcn{``利益''} (benefits) in the summary is not related to the source content. Thus, it has to be remembered by the model, together with the source content. However, this correspondence is spurious, because the word \zhcn{``利益''} is not related to any word in the source content. In the following, we refer to this problem as \textbf{Spurious Correspondence} caused by the semantically inconsistent data. 

In this work, we aim to alleviate the impact of the semantic inconsistency of the current dataset. Based on the sequence-to-sequence model, we propose a regularization method to heuristically show down the learning of the spurious correspondence, so that the unrelated information in the dataset is less represented by the model. We incorporate a new soft training target to achieve this goal. For each output time in training, in addition to the gold reference word, the current output also targets at a softened output word distribution that regularizes the current output word distribution. In this way, a more robust correspondence of the source content and the output words can be learned, and potentially, the output summary will be more semantically consistent.

To obtain the softened output word distribution, we propose two methods based on the sequence-to-sequence model. 
\begin{itemize}
\item The first one uses the output layer of the decoder to generate the distribution but with a higher temperature when using softmax normalization. It keeps the relative order of the possible output words but guides the model to keep a smaller discriminative margin. For spurious correspondence, across different examples, the output distribution is more likely to be different, so no effective discriminative margin will be established. For true correspondence, across different examples, the output distribution is more likely to be the same, so a margin can be gradually established.
\item The second one introduces an additional output layer to generate the distribution. Analogous to multi-task learning, the additional output layer provides an alternative view of the data, so that it can regularize the output distribution more effectively. Because the additional output layer differs from the original one in that the less stable information, i.e., the spurious correspondence learned by the model itself, is represented differently. Besides, the relative order can also be regularized in this method.
\end{itemize}
More detailed explanation is introduced in Section \ref{sec:method}.

Another problem for abstractive text summarization is that the system summary cannot be easily evaluated automatically. ROUGE \cite{rouge} is widely used for summarization evaluation. However, as ROUGE is designed for extractive text summarization, it cannot deal with summary paraphrasing in abstractive text summarization. Besides, as ROUGE is based on the reference, it requires high-quality reference summary for a reasonable evaluation, which is also lacking in the existing dataset for Chinese social media text summarization. We argue that for proper evaluation of text generation task, human evaluation cannot be avoided. We propose a simple and practical human evaluation for evaluating text summarization, where the summary is evaluated against the source content instead of the reference. It handles both of the problems of paraphrasing and lack of high-quality reference.

The contributions of this work are summarized as follows:
\begin{itemize}
\item We propose an approach to regularize the output word distribution, so that the semantic inconsistency, e.g. words not related to the source content, exhibited in the training data is underrepresented in the model. We add a cross-entropy based regularization term to the overall loss. We also propose two methods to obtain the soft target distribution for regularization. The results demonstrate the effectiveness of the proposed approach, which outperforms almost all the existing systems. In particular, the semantic consistency is improved by 4\% in terms of human evaluation. We also conduct analysis to examine the effect of the proposed method on the output summaries and the output label distributions, showing that the improved consistency results from the regularized output distribution.

\item We propose a simple human evaluation method to assess the semantic consistency of the generated summary with the source content. Such kind of evaluation is absent in the existing work of text summarization. In the proposed human evaluation, the summary is evaluated against the source content other than the reference summary, so that it can better measure the consistency of the generated summary and the source content when high-quality reference is not available.
\end{itemize}

\section{Proposed Method\label{sec:method}}

Base on the fact that the spurious correspondence is not stable and its realization in the model is prone to change, we propose to alleviate the issue heuristically by regularization. We use the cross-entropy with an annealed output distribution as the regularization term in the loss so that the little fluctuation in the distribution will be depressed and more robust and stable correspondence will be learned. By correspondence, we mean the relation between (a) the current output, and (b) the source content and the partially generated output. Furthermore, we propose to use an additional output layer to generate the annealed output distribution. Due to the same fact, the two output layers will differ more in the words that superficially co-occur, so that the output distribution can be better regularized.


\begin{figure}[htbp]
\centering
\includegraphics[width=0.8\linewidth]{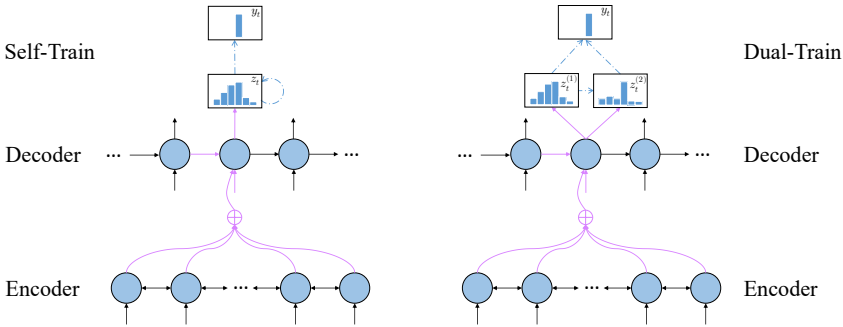}
\caption{Illustration of the proposed methods. Left: Self-Train. Right: Dual-Train. 
\label{fig:model}}

\end{figure}

\subsection{Regularizing the Neural Network with Annealed Distribution}

Typically, in the training of the sequence-to-sequence model, only the one-hot hard target is used in the cross-entropy based loss function. For an example in the training set, the loss of an output vector is\footnote{For the convenience of description, we omit the related trainable parameters of the model.}
\begin{equation}
	J(\vib{z}) = \sum_{i=1}^{M} - y_i \log z_i
\end{equation}
where $\vib{z}$ is the output vector, $\vib{y}$ is the one-hot hard target vector, and $M$ is the number of labels. However, as $\vib{y}$ is the one-hot vector, all the elements are zero except the one representing the correct label. Hence, the loss becomes
\begin{equation}
	J(\vib{z}) = - \log z_l
\end{equation}
where $l$ is the index of the correct label.
The loss is then summed over the output sentences and across the minibatch and used as the source error signal in the backpropagation.

The hard target could cause several problems in the training. Soft training methods try to use a soft target distribution to provide a generalized error signal to the training. For the summarization task, a straight-forward way would be to use the current output vector as the soft target, which contains the knowledge learned by the current model, i.e., the correspondence of the source content and the current output word:
\begin{equation}
J(\vib{z}) = \sum_{i=1}^{M} - z_i \log z_i \label{eq:self-learning}
\end{equation}
Then, the two losses are combined as the new loss function:
\begin{equation}
	J(\vib{z}) = - \log z_l - \alpha \sum_{i=1}^{M} z_i \log z_i
\end{equation}
where $l$ is the index of the true label and $\alpha$ is the strength of the soft training loss. We refer to this approach as \textbf{Self-Train} (The left part of Figure \ref{fig:model}).

The output of the model can be seen as a refined supervisory signal for the learning of the model. The added loss promotes the learning of more stable correspondence. The output not only learns from the one-hot distribution but also the distribution generated by the model itself. However, during the training, the output of the neural network can become too close to the one-hot distribution. To solve this, we make the soft target the soften output distribution. We apply the softmax with temperature $\tau$, which is computed by\footnote{It uses simplified notation, where $z_i$ and $z_j$ should be the unnormalized output, i.e., the output before the softmax operation.} 
\begin{equation}
z'_i = \frac{\exp(z_i/\tau)}{\sum_{j=1}^{M} \exp(z_j/\tau)}, \quad i =1,2,\ldots,M \label{eq:softmax-temperature}
\end{equation}
This transformation keeps the relative order of the labels, and a higher temperature will make the output distributed more evenly.

The key motivation is that if the model is still not confident how to generate the current output word under the supervision of the reference summary, it means the correspondence can be spurious and the reference output is unlikely to be concluded from the source content. It makes no sense to force the model to learn such correspondence. The regularization follows that motivation, and in such case, the error signal will be less significant compared to the one-hot target. In the case where the model is extremely confident how to generate the current output, the annealed distribution will resemble the one-hot target. Thus, the regularization is not effective. In all, we make use of the model itself to identify the spurious correspondence and then regularize the output distribution accordingly.


\subsection{Dual Output Layers}

However, the aforementioned method tries to regularize the output word distribution based on what it has already learned. The relative order of the output words is kept. The self-dependency may not be desirable for regularization. It may be better if more correspondence that is spurious can be identified.


In this paper, we further propose to obtain the soft target from a different view of the model, so that different knowledge of the dataset can be used to mitigate the overfitting problem. An additional output layer is introduced to generate the soft target. The two output layers share the same hidden representation but have independent parameters. They could learn different knowledge of the data. We refer to this approach as \textbf{Dual-Train}. For clarity, the original output layer is denoted by $o^{(1)}$ and the new output layer $o^{(2)}$. Their outputs are denoted by $\vib{z^{(1)}}$ and $\vib{z^{(2)}}$, respectively.

The output layer $o^{(1)}$ acts as the original output layer. We apply soft training using the output from $o^{(2)}$ to this output layer to increase its ability of generalization. Suppose the correct label is $l$. The target of the output $\vib{z^{(1)}}$ includes both the one-hot distribution and the distribution generated from $o^{(2)}$:
\begin{equation}
J(\vib{z^{(1)}}) = -\log z_l^{(1)} 
                   -\alpha \sum_{i=1}^{M} z_i^{(2)\prime} \log z_i^{(1)}
\end{equation}

The new output layer $o^{(2)}$ is trained normally using the originally hard target. This output layer is not used in the prediction, and its only purpose is to generate the soft target to facilitate the soft training of $o^{(1)}$. Suppose the correct label is $l$. The target of the output $\vib{z^{(2)}}$ includes only the one-hot distribution:
\begin{equation}
\begin{split}
J(\vib{z^{(2)}}) = &-\log z_l^{(2)} 
\end{split}
\end{equation}
Because of the random initialization of the parameters in the output layers, $o^{(1)}$ and $o^{(2)}$ could learn different things. The diversified knowledge is helpful when dealing with the spurious correspondence in the data. 
It can also be seen as an online kind of ensemble methods. Several different instances of the same model are softly aggregated into one to make classification. The right part of Figure~\ref{fig:model} shows the architecture of the proposed Dual-Train method. 

\section{Experiments}

We evaluate the proposed approach on the Chinese social media text summarization task, based on the sequence-to-sequence model. We also analyze the output text and the output label distribution of the models, showing the power of the proposed approach. Finally, we show the cases where the correspondences learned by the proposed approach are still problematic, which can be explained based on the approach we adopt.

\subsection{Dataset}

Large-Scale Chinese Short Text Summarization Dataset (LCSTS) is constructed by \cite{lcsts}. The dataset consists of more than 2.4 million text-summary pairs in total, constructed from a famous Chinese social media microblogging service Weibo\footnote{\url{http://weibo.com}}. The whole dataset is split into three parts, with 2,400,591 pairs in PART I for training, 10,666 pairs in PART II for validation, and 1,106 pairs in PART III for testing. The authors of the dataset have manually annotated the relevance scores, ranging from 1 to 5, of the text-summary pairs in PART II and PART III. They suggested that only pairs with scores no less than three should be used for evaluation, which leaves 8,685 pairs in PART II, and 725 pairs in PART III. From the statistics of the PART II and PART III, we can see that more than 20\% of the pairs are dropped to maintain semantic quality. It indicates that the training set, which has not been manually annotated and checked, contains a huge quantity of unrelated text-summary pairs. 

\subsection{Experimental Settings}

We use the sequence-to-sequence model \cite{seq2seq} with attention \cite{attention,stanfordattention,mapattention,supervisedattention} as the \textbf{Baseline}. Both the encoder and decoder are based on the single layer LSTM \cite{Hochreiter1997}. The word embedding size is 400, and the hidden state size of the LSTM unit is 500. We conduct experiments on the word level. To convert the character sequences into word sequences, we use Jieba\footnote{\url{https://pypi.python.org/pypi/jieba/}} to segment the words, the same with the existing work \cite{lcsts,copynet}.

Self-Train and Dual-Train are implemented based on the baseline model, with two more hyper-parameters, the temperature $\tau$ and the soft training strength $\alpha$. We use a very simple setting for all tasks, and set $\tau=2$, $\alpha=1$. We pre-train the model without applying the soft training objective for 5 epochs out of total 10 epochs.  
We use the Adam optimizer \cite{Kingma2014} for all the tasks, using the default settings with $\alpha=0.001$, $\beta_0 = 0.999$, and $\beta_1=0.9$. In testing, we use beam search to generate the summaries, and the beam size is set to 5. We report the test results at the epoch that achieves the best score on the development set.

\subsection{Evaluation Protocol}

For text summarization, a common automatic evaluation method is ROUGE \cite{rouge}. The generated summary is evaluated against the reference summary, based on unigram recall (ROUGE-1), bigram recall (ROUGE-2), and recall of longest common subsequence (ROUGE-L). To facilitate comparison with the existing systems, we adopt ROUGE as the automatic evaluation method. The ROUGE is calculated on the character level, following the previous work \cite{lcsts}.

However, for abstractive text summarization, the ROUGE is sub-optimal, and cannot assess the semantic consistency between the summary and the source content, especially when there is only one reference for a piece of text. The reason is that the same content may be expressed in different ways with different focuses. Simple word match cannot recognize the paraphrasing. It is the case for all of the existing large-scale datasets. Besides, as aforementioned, ROUGE is calculated on the character level in Chinese text summarization, making the metrics favor the models on the character level in practice. In Chinese, a word is the smallest semantic element that can be uttered in isolation, not a character. In the extreme case, the generated text could be completely intelligible, but the characters could still match. In theory, calculating ROUGE metrics on the word level could alleviate the problem. However, word segmentation is also a non-trivial task for Chinese. There are many kinds of segmentation rules, which will produce different ROUGE scores. We argue that it is not acceptable to introduce additional systematic bias in automatic evaluations, and automatic evaluation for semantically related tasks can only serve as a reference.

To avoid the deficiencies, we propose a simple human evaluation method to assess the semantic consistency. Each summary candidate is evaluated against the text rather than the reference. If the candidate is irrelevant or incorrect to the text, or the candidate is not understandable, the candidate is labeled \textit{bad}. Otherwise, the candidate is labeled \textit{good}. Then, we can get an accuracy of the good summaries. The proposed evaluation is very simple and straight-forward. It focuses on the relevance between the summary and the text. The semantic consistency should be the major consideration when putting the text summarization methods into practice, but the current automatic methods cannot judge properly. For detailed guidelines in human evaluation, please refer to Appendix \ref{sec:he-std}.

In the human evaluation, the text-summary pairs are dispatched to two human annotators who are native speakers of Chinese. As in our setting the summary is evaluated against the reference, the number of the pairs needs to be manually evaluated is four times the number of the pairs in the test set, because we need to compare four systems in total. To decrease the workload and get a hint about the annotation quality at the same time, we adopt the following procedure. We first randomly select 100 pairs in the validation set for the two human annotators to evaluate. Each pair is annotated twice, and the inter-annotator agreement is checked. We find that under the protocol, the inter-annotator agreement is quite high. In the evaluation of the test set, a pair is only annotated once to accelerate evaluation. To further maintain consistency, summaries of the same source content will not be distributed to different annotators.

\subsection{Experimental Results}

\begin{table}[htbp]

\centering

\caption{Results of the human evaluation, showing how many summaries are semantically consistent with their source content. The generated summary is evaluated directly against the source content.\label{tab:res-hu}}

	\begin{tabular}{l c c c }
		\hline
		\bf Methods & \bf \# Good & \bf \# Total & \bf Accuracy   \\ 
        \hline
        Reference & 673  & \multirow{4}{*}{725} & 92.8\% \\ 
		Baseline  & 360 & &  49.6\% \\
        Self-Train & 316 & & 43.6\%\\
        Dual-Train & \bf 389  & & \bf 53.6\% \\ 
        \hline
	\end{tabular}

\end{table}

First, we show the results for human evaluation, which focuses on the semantic consistency of the summary with its source content. We evaluate the systems implemented by us as well as the reference. We cannot conduct human evaluations for the existing systems from other work, because the output summaries needed are not available for us. Besides, the baseline system we implemented is very competitive in terms of ROUGE and achieves better performance than almost all the existing systems. The results are listed in Table~\ref{tab:res-hu}. It is surprising to see that the accuracy of the reference summaries does not reach 100\%. It means that the test set still contains text-summary pairs of poor quality even after removing the pairs with relevance scores lower than 3 as suggested by the authors of the dataset. As we can see, Dual-Train improves the accuracy by 4\%. Due to the rigorous definition of being good, the results mean that 4\% more of the summaries are semantically consistent with their source content. However, Self-Train has a performance drop compared to the baseline. After investigating its generated summaries, we find that the major reason is that the generated summaries are not grammatically complete and often stop too early, although the generated part is indeed more related to the source content. Because the definition of being good, the improved relevance does not make up the loss on intelligibility.

\begin{table}[htbp]
\centering

\caption{Comparisons with the existing models in terms of ROUGE metrics. \label{tab:res}}

	\begin{tabular}{l c c c}
		\hline
		\textbf{Methods} & \textbf{ROUGE-1} &\textbf{ROUGE-2} &  \textbf{ROUGE-L}   \\ 
        \hline
		RNN-context \cite{lcsts} & 29.9 & 17.4 & 27.2 \\
		SRB \cite{MaEA2017}& 33.3 & 20.0 & 30.1 \\
		CopyNet \cite{copynet} & 35.0 & 22.3 & 32.0 \\ 
		RNN-distract \cite{distraction} & 35.2 & 22.6 & 32.5 \\
		DRGD \cite{DRGD} & 37.0 & 24.1 &  34.2  \\  
        \hline
		Baseline (Ours) & 35.3 & 23.4& 33.0 \\
        Self-Train (Ours) &35.3 &23.3 & 32.6 \\
        Dual-Train (Ours) & 36.2  & 24.3 & 33.8 \\ 
        \hline
		
	\end{tabular}

\end{table}

Then, we compare the automatic evaluation results in Table~\ref{tab:res}. As we can see, only applying soft training without adaptation (Self-Train) hurts the performance. With the additional output layer (Dual-Train), the performance can be greatly improved over the baseline. Moreover, with the proposed method the simple baseline model is second to the best compared with the state-of-the-art models and even surpasses in ROUGE-2. It is promising that applying the proposed method to the state-of-the-art model could also improve its performance. 

The automatic evaluation is done on the original test set to facilitate comparison with existing work. However, a more reasonable setting would be to exclude the 52 test instances that are found bad in the human evaluation, because the quality of the automatic evaluation depends on the reference summary. As the existing methods do not provide their test output, it is a non-trivial task to reproduce all their results of the same reported performance. Nonetheless, it does not change the fact that ROUGE cannot handle the issues in abstractive text summarization properly.

\subsection{Experimental Analysis}

To examine the effect of the proposed method and reveal how the proposed method improves the consistency, we compare the output of the baseline with Dual-Train, based on both the output text and the output label distribution. We also conduct error analysis to discover room for improvements.

\subsubsection{Analysis of the Output Text}

\begin{table}[htbp]
\centering

\caption{Examples of the summaries generated by the baseline and Dual-Train from the test set. As we can see, the summaries generated by the proposed are much better than the ones generated by the baseline, and even more informative and precise than the references. \label{tab:ex-res}}

\setlength{\tabcolsep}{3pt}
\scriptsize

\begin{tabular}{|p{.95\linewidth}|}
\hline
\textbf{Short Text:}
\zhcn{中国铁路总公司消息，自2015年1月5日起，自行车不能进站乘车了。骑友大呼难以接受。这部分希望下车就能踏上骑游旅程的旅客，只能先办理托运业务，可咨询12306客服电话，就近提前办理。运费每公斤价格根据运输里程不同而不同。}\\
China Railway news: Starting from January 5, 2015, bikes cannot be carried onto the platform and the train. Riders voice complaints. The travelers, who wish that as soon as they get off the train, their bike ride can begin, have to check their bikes in before getting on board; they can consult the customer service at 12306, and check in at the nearest service station in advance. Shipping costs per kilogram depend on the shipping mileage. \\
\hline
\textbf{Reference:} \zhcn{自行车不能\cjkhl{带上}火车} 
Bikes cannot be carried onto the train \\
\textbf{Baseline:} \zhcn{自行车：自行车不能\cjkhl{停}，你怎么看？} 
Bikes: Bikes will not stop; What do you think of it?\\
\textbf{Proposal:} \zhcn{铁路总公司：自行车不能\cjkhl{进站}} 
China Railway: Bikes cannot be carried onto the platform\\
\hline
\hline
\textbf{Short Text:}
\zhcn{11日下午，中共中央政治局常委、中央书记处书记刘云山登门看望了国家最高科技奖获得者于敏、张存浩。刘云山指出，广大科技工作者要学习老一辈科学家求真务实的钻研精神，淡泊名利、潜心科研，努力创造更多一流科研成果。}\\
On the afternoon of the 11th, Liu Yunshan, member of the Standing Committee of the Political Bureau of the CPC Central Committee and secretary of the Secretariat of the CPC Central Committee, paid a visit to Yu Min and Zhang Cunhao, the recipients of the State Preeminent Science and Technology Award. Liu Yunshan pointed out that scientists and technologists should study the pragmatic research spirit of the older generation, be indifferent to fame and fortune, devote themselves to scientific research, and strive to create more first-rate scientific research achievements.\\
\hline
\textbf{Reference:} \zhcn{刘云山看望著名科技\cjkhl{专家}} 
Liu Yunshan paid a visit to prominent science and technology experts\\
\textbf{Baseline:} \zhcn{刘云山：科技\cjkhl{钻研}钻研} 
Liu Yunshan: Science and technology research research\\
\textbf{Proposal:} \zhcn{刘云山：科技\cjkhl{工作者}要学习老一辈科学家的钻研精神} 
Liu Yunshan: Scientists and technologists should study the pragmatic research spirit of the older generation\\
\hline
\hline
\textbf{Short Text:}
\zhcn{9月3日，总部位于日内瓦的世界经济论坛发布了《2014-2015年全球竞争力报告》，瑞士连续六年位居榜首，成为全球最具竞争力的国家，新加坡和美国分列第二位和第三位。中国排名第28位，在金砖国家中排名最高。
}\\
On September 3, the Geneva-based World Economic Forum released the Global Competitiveness Report 2014-2015. Switzerland ranks first for six years in a row, becoming the most competitive country in the world, with Singapore and the United States taking the second place and the third place. China ranks the 28th, the highest in the BRIC countries. \\
\hline
\textbf{Reference:} \zhcn{全球竞争力排行榜中国\cjkhl{居}28位居金砖国家首位} 
In the Global Competitiveness Rankings, China ranks the 28th, the highest in the BRIC countries \\
\textbf{Baseline:} \zhcn{全球竞争力报告：中国\cjkhl{居}全球第八} 
Global Competitiveness Report: China ranks the eighth in the world  \\
\textbf{Proposal:} \zhcn{全球竞争力排名：瑞士居榜首中国\cjkhl{位列}第28} 
Global Competitiveness Report: Switzerland takes the first place; China ranks the 28th  \\
\hline
\end{tabular}

\end{table}

To gain a better understanding of the results, we analyze the summaries generated by the baseline model and our proposed model. Some of the summaries are listed in Table~\ref{tab:ex-res}. As shown in the table, the summaries generated by the proposed method are much better than the baseline, and we believe they are more precise and informative than the references.

In the first one, the baseline system generates a grammatical but unrelated summary, while the proposed method generates a more informative summary. In the second one, the baseline system generates a related but ungrammatical summary, while the proposed method generates a summary related to the source content but different from the reference. We believe the generated summary is actually better than the reference because the focus of the visit is not the event itself but its purpose. In the third one, the baseline system generates a related and grammatical summary, but the facts stated are completely incorrect. The summary generated by the proposed method is more comprehensive than the reference, while the reference only includes the facts in the last sentence of the source content.

In short, the generated summary of the proposed method is more consistent with the source content. It also exhibits the necessity of the proposed human evaluation. Because when the generated summary is evaluated against the reference, it may seem redundant or wrong, but it is actually true to the source content. While it is arguable that the generated summary is better than the reference, there is no doubt that the generated summary of the proposed method is better than the baseline. However, the improvement cannot be properly shown by the existing evaluation methods.

Furthermore, the examples suggest that the proposed method does learn better correspondence. The highlighted words in each example in Table~\ref{tab:ex-res} share almost the same previous words. However, in the first one, the baseline considers ``\zhcn{停}'' (stop) as the most related words, which is a sign of noisy word relations learned from other training examples, while the proposed method generates ``\zhcn{进站}'' (to the platform), which is more related to what a human thinks. It is the same with the second example, where a human selects ``\zhcn{专家}'' (expert) and Dual-Train selects ``\zhcn{工作者}'' (worker), while the baseline selects ``\zhcn{钻研}'' (research) and fails to generate a grammatical sentence later. In the third one, the reference and the baseline use the same word, while Dual-Train chooses a word of the same meaning. It can be concluded that Dual-Train indeed learns better word relations that could generalize to the test set, and good word relations can guide the decoder to generate semantically consistent summaries.

\subsubsection{Analysis of the Output Label Distribution}

\begin{table}[htbp]
    \centering
     \caption{Examples of the labels and their top 4 most related labels. The highlighted words indicate problematic word relations. The baseline system encodes semantically unrelated words into the word relations, while the proposed method learns the relatedness more precisely and robustly.}
    \label{tab:heatmap-textsum}
    \footnotesize
    \setlength{\tabcolsep}{2pt}
       \begin{tabular}{l l}
      \hline
      Label (Word)	&	Top 4 Most Related Labels \\
      \hline
      \zhcn{我国}				&	\textbf{Proposal:} 		\ex{中国}{China}, \ex{全国}{nationwide}, \ex{国家}{nation}, \ex{国内}{domestic}\\
      (our country)	&	\textbf{Baseline:} 		\ex{中国}{China}, \ex{全国}{nationwide}, \ex{国内}{domestic}, \exh{国务院}{the State Council}\\ 
      \hline
      
      \zhcn{旧车}				&	\textbf{Proposal:} 		\ex{机动车}{motor vehicle}, \ex{汽车}{automobile}, \exh{北京}{Beijing}, \ex{司机}{driver}\\
      (old car)  			&	\textbf{Baseline:} 		\exh{限牌}{restricted license}, \exh{北京}{Beijing}, \ex{车辆}{vehicle}, \exh{补助}{subsidy}\\ 
      \hline
      
      \zhcn{一名}			&	\textbf{Proposal:} 	   \ex{1}{one}, \ex{一}{one}, \ex{一人}{one \textit{REN}}, \ex{一个}{one \textit{GE}}\\
      (one \textit{MING})				&	\textbf{Baseline:} 		\ex{一}{one}, \ex{1}{one}, \exh{中国}{China}, \ex{一个}{one \textit{GE}}\\ 
      \hline
      
      \zhcn{图像}				&	\textbf{Proposal:} 		\ex{图}{figure}, \ex{照片}{photo}, \ex{影像}{image}, \ex{图片}{picture}\\
      (image)		& \textbf{Baseline:} 		\ex{照片}{photo}, \ex{图}{figure}, \exh{和}{and}, \exh{的}{\textit{DE}}\\ 
      \hline
      
      \zhcn{tesla}				&	\textbf{Proposal:} 		\ex{特斯拉}{tesla}, \ex{汽车}{automobile}, \ex{电动车}{electrombile}, \ex{电动汽车}{electric vehicle}\\
      (tesla)  &	\textbf{Baseline:} 		\ex{特斯拉}{tesla}, \ex{汽车}{automobile}, \exh{美国}{USA}, \exh{互联网}{Internet}\\ 
      \hline
      
      \zhcn{期货业}			&	\textbf{Proposal:} 		\ex{期货}{futures}, \exh{改革}{reform}, \ex{证券}{bond}, \ex{企业}{enterprise}\\
      (futures industry)	&	\textbf{Baseline:} 		\ex{期货}{futures}, \exh{中国}{China}, \exh{2013}{2013}, \exh{创新}{innovation}\\ 
      \hline
      
      \zhcn{雨夹雪}			&	\textbf{Proposal:} \ex{下雪}{snow}, \ex{雨雪}{sleet}, \ex{降雪}{snowfall}, \ex{大风}{strong wind}  \\
      (sleet) 					&	\textbf{Baseline:} \ex{雨雪}{sleet}, \exh{有}{exist}, \ex{降雪}{snowfall}, \ex{初雪}{first snowfall}\\
      \hline
      
      \zhcn{多长时间}			&	\textbf{Proposal:} \ex{多少}{how many}, \ex{多久}{how long}, \exh{的}{\textit{DE}}, \ex{时间}{time}  \\
      (how long) 					&	\textbf{Baseline:} \exh{吗}{\textit{MA}}, \exh{197}{197}, \exh{的}{\textit{DE}}, \exh{知道}{know}\\
      \hline
      
      \zhcn{沙尘暴}			&	\textbf{Proposal:} \ex{大风}{windstorm}, \ex{沙尘}{sand and dust}, \exh{新疆}{Xinjiang}, \exh{美国}{USA}  \\
      (sandstorm) 					&	\textbf{Baseline:} \ex{沙尘}{sand and dust}, \exh{暴雪}{snowstorm}, \exh{暴雨}{rainstorm}, \exh{短时}{short time}\\
      \hline
      
       \zhcn{乘警}			&	\textbf{Proposal:} \ex{铁警}{railway police}, \ex{民警}{police}, \ex{列车}{train}, \ex{铁路}{railway}  \\
      (railway police) 					&	\textbf{Baseline:} \exh{救人}{rescue somebody}, \ex{铁警}{railway police}, \ex{警察}{police}, \exh{实习}{internship}\\
      \hline
   \end{tabular}

\end{table}

To show why the generated text of the proposed method is more related to the source content, we further analyze the label distribution, i.e., the word distribution, generated by the (first) output layer, from which the output word is selected. To illustrate the relationship, we calculate a representation for each word based on the label distributions. Each representation is associated with a specific label (word), denoted by $l$, and each dimension $i$ shows how likely the label indexed by $i$ will be generated instead of the label $l$. To get such representation, we run the model on the training set and get the output vectors in the decoder, which are then averaged with respect to their corresponding labels to form a representation. We can obtain the most related words of a word by simply selecting the highest values from its representation. Table~\ref{tab:heatmap-textsum} lists some of the labels and the top 4 labels that are most likely to replace each of the labels. It is a hint about the correspondence learned by the model.

From the results, it can be observed that Dual-Train learns the better semantic relevance of a word compared to the baseline because the spurious word correspondence is alleviated by regularization. For example, the possible substitutes of the word ``\zhcn{多长时间}'' (how long) considered by Dual-Train include ``\zhcn{多少}'' (how many), ``\zhcn{多久}'' (how long) and ``\zhcn{时间}'' (time). However, the relatedness is learned poorly in the baseline, as there is ``\zhcn{知道}'' (know), a number, and two particles in the possible substitutes considered by the baseline. Another representative example is the word ``\zhcn{图像}'' (image), where the baseline also includes two particles in its most related words. 

The phenomenon shows that the baseline suffers from spurious correspondence in the data, and learns noisy and harmful relations, which rely too much on the co-occurrence. In contrast, the proposed method can capture more stable semantic relatedness of the words. For text summarization, grouping the words that are in the same topic together can help the model to generate sentences that are more coherent and can improve the quality of the summarization and the relevance to the source content.

Although the proposed method resolves a large number of the noisy word relations, there are still cases that the less related words are not eliminated. For example, the top 4 most similar words of ``\zhcn{期货业}'' (futures industry) from the proposed method include ``\zhcn{改革}'' (reform). It is more related than ``2013'' from the baseline, but it can still be harmful to text summarization. The problem could arise from the fact that words as ``\zhcn{期货业}'' rarely occur in the training data, and their relatedness is not reflected in the data. Another issue is that there are some particles, e.g., ``\zhcn{的}'' (DE) in the most related words. A possible explanation is that particles show up too often in the contexts of the word, and it is hard for the models to distinguish them from the real semantically-related words. As our proposed approach is based on regularization of the less common correspondence, it is reasonable that such kind of relation cannot be eliminated. The first case can be categorized into data sparsity, which usually needs the aid of knowledge bases to solve. The second case is due to the characteristics of natural language. However, as such words are often closed class words, the case can be resolved by manually restricting the relatedness of these words.

\section{Related Work\label{sec:relatedwork}}

Related work includes efforts on designing models for the Chinese social media text summarization task and the efforts on obtaining soft training target for supervised learning.

\subsection{Systems for Chinese Social Media Text Summarization}

The Large-Scale Chinese Short Text Summarization dataset was proposed by \cite{lcsts}. Along with the datasets, \cite{lcsts} also proposed two systems to solve the task, namely \textit{RNN} and \textit{RNN-context}. They were two sequence-to-sequence based models with GRU as the encoder and the decoder. The difference between them was that RNN-context had attention mechanism while RNN did not. They conducted experiments both on the character level and on the word level. \textit{RNN-distract}~\cite{distraction} was a distraction-based neural model, where the attention mechanism focused on different parts of the source content.
\textit{CopyNet}~\cite{copynet} incorporated a copy mechanism to allow part of the generated summary to be copied from the source content. The copy mechanism also explained that the results of their word-level model were better than the results of their character-level model. \textit{SRB}~\cite{MaEA2017} was a sequence-to-sequence based neural model to improve the semantic relevance between the input text and the output summary. \textit{DRGD}~\cite{DRGD} was a deep recurrent generative decoder model, combining the decoder with a variational autoencoder.

\subsection{Methods for Obtaining Soft Training Target}

Soft target aims to refine the supervisory signal in supervised learning. Related work includes soft target for traditional learning algorithms and model distillation for deep learning algorithms.

The soft label methods are typically for binary classification \cite{NguyenVH14}, where the human annotators not only assign a label for an example but also give information on how confident they are regarding the annotation. 
The main difference from our method is that the soft label methods require additional annotation information (e.g., the confidence information of the annotated labels) of the training data, which is costly in the text summarization task.

There have also been prior studies on model distillation in deep learning that distills big models into a smaller one.
Model distillation \cite{distillation} combined different instances of the same model into a single one. It used the output distributions of the previously trained models as the soft target distribution to train a new model. 
A similar work to model distillation is the soft-target regularization method~\cite{Aghajanyan16} for image classification. Instead of using the outputs of other instances, it used an exponential average of the past label distributions of the current instance as the soft target distribution.  
The proposed method is different compared with the existing model distillation methods, in that the proposed method does not require additional models or additional space to record the past soft label distributions. The existing methods are not suitable for text summarization tasks, because the training of an additional model is costly, and the additional space is huge due to the massive number of data. The proposed method uses its current state as the soft target distribution and eliminates the need to train additional models or to store the history information.

\section{Conclusions}
We propose a regularization approach for the sequence-to-sequence model on the Chinese social media summarization task. In the proposed approach, we use a cross-entropy based regularization term to make the model neglect the possible unrelated words. 
We propose two methods for obtaining the soft output word distribution used in the regularization, of which Dual-Train proves to be more effective.
Experimental results show that the proposed method can improve the semantic consistency by 4\% in terms of human evaluation. As shown by the analysis, the proposed method achieves the improvements by eliminating the less semantically-related word correspondence.

The proposed human evaluation method is effective and efficient in judging the semantic consistency, which is absent in previous work but is crucial in the accurate evaluation of the text summarization systems. The proposed metric is simple to conduct and easy to interpret. It also provides an insight on how practicable the existing systems are in the real-world scenario.

\appendix

\section{Standard for Human Evaluation\label{sec:he-std}}

\begin{table}[htbp]
\centering
\caption{Examples for each case in the human evaluation. There are three rules to be examined in order. If one rule is not met, the following rules do not need to be checked. The procedure leaves us four specific cases in total, of which three is bad, and one is good. \label{ex:he-rule} }

\scriptsize
\begin{tabular}{ | p{.95\textwidth} |  }
\hline
\multicolumn{1}{|c|}{\bf Bad Fluency} \\
\hline
\textbf{Source:} \zhcn{今年以来，多家券商都在``找婆家''。7月8日，齐鲁证券4亿股权在北京金融资产交易所挂牌转让，加上目前正在四大产权交易所挂牌转让的世纪证券、申银万国、云南证券等，至少4家券商股权亮相于各地产权交易所。} \\
Since the beginning of this year, a number of brokers have been ``finding their husbands''. On July 8th, 400 million shares of Qilu Securities were listed on the Beijing Financial Asset Exchange, together with Century Securities, Shenyin Wanguo, and Yunnan Securities, which are currently listed on the four major property rights exchanges, and at least four stockholders’ equity was unveiled around the country.\\
\textbf{Summary:} \zhcn{齐鲁证券4亿股权“找”} Qilu Securities 400 million shares ``find''\\
\hline \hline
\multicolumn{1}{|c|}{\bf Good Fluency, Bad Relatedness} \\
\hline
\textbf{Source:} \zhcn{2013年8月，小米公司估值已经超过100亿美元。这主要是因为小米不是以一家硬件公司估值的公司。在智能终端设备产业中，硬件被视为产业链下游的公司，用宏碁创始人施政荣的话说就是``处于微笑曲线的底部''。} \\
In August 2013, Xiaomi's valuation has exceeded US\$10 billion. This is mainly because Xiaomi is not a company that is valued by a hardware company. In the smart terminal equipment industry, the hardware company is regarded as a company in the downstream of the industrial chain, and it is ``in the bottom of a smile curve'' in the words of Acer's founder Shi Zhengrong.\\
\textbf{Summary:} \zhcn{小米渠道保卫战：雷军多措施打击黄牛} Xiaomi's defense in marketing channels: Lei Jun combats the scalpers with multiple measures \\
\hline \hline
\multicolumn{1}{|c|}{\bf Good Fluency and Relatedness, Bad Faithfulness} \\
\hline
\textbf{Source:} \zhcn{田同生：62岁，42次跑马。央视解说员于嘉：卖了车，每天跑7公里上班。联想副总裁魏江雷：跑步可以放空大脑；地产CEO刘爱明：跑马拉松始终有一个情形，会刺激我：无论跑多快，都能看见明显比你年长的人跑在你前面。}\\
Tian Tongsheng: 62 years old, 42 times marathon. CCTV narrator Yu Jia: Selling a car, he runs 7 kilometers a day to work. Lenovo vice president Wei Jianglei: Running can empty the brain; real estate CEO Liu Aiming: Running a marathon always has a situation that will stimulate me: No matter how fast you run, you can see people who are obviously older than you running in front of you. \\
\textbf{Summary:} \zhcn{柳传志：62岁42次每天跑7公里上班} Liu Chuanzhi: 62 years old, 42 times marathon, he runs 7 kilometers a day to work.\\
\hline \hline
\multicolumn{1}{|c|}{\bf Good Fluency, Relatedness, and Faithfulness} \\
\hline
\textbf{Source:} \zhcn{成都市软件和信息技术服务业近年来一直保持快速增长势头，稳居中西部城市之首，已成为我国西部``硅谷''。《2013年度成都市软件和信息技术服务产业发展报告》日前发布}\\
Chengdu's software and information technology service industry has maintained a rapid growth momentum in recent years, ranking first in the Midwest cities, and it has become the ``Silicon Valley'' in western China. ``2013 Chengdu Software and Information Technology Service Industry Development Report'' has been released.\\
\textbf{Summary:} \zhcn{成都倾力打造西部``硅谷''} Chengdu's efforts to build west ``Silicon Valley'' \\
\hline
\end{tabular}

\end{table}

For human evaluation, the annotators are asked to evaluate the summary against the source content based on the goodness of the summary. If the summary is not understandable, relevant or correct according to the source content, the summary is considered bad. More concretely, the annotators are asked to examine the following aspects to determine whether the summary is \textit{good}: 
\begin{enumerate}

\item \textbf{Fluency.} If the summary itself cannot be understood, the summary is not good. This is not judged by grammatical correctness but by checking if major semantic roles, such as the predicate, the agent, and the experiencer, are missing. It allows the summary where certain particles (e.g., \zhcn{``的''} or aspect markers (e.g., \zhcn{``了''}) are missing but the content can still be understood. Another common issue is that the summary may have repeated words. Our view on this is that if the repetition does not affect the intelligibility of the summary, we still treat the summary as good. 

\item \textbf{Relatedness.} If it is impossible to decide whether the summary is correct or wrong according to the source content, the summary is bad. By definition, the nonsense is also ruled out.

\item \textbf{Faithfulness.} If the summary is not correct according to the source content, the summary is labeled bad. We also treat the summary that is relevant and correct according to the source content but is too general as bad.
\end{enumerate}

If a rule is not met, the summary is labeled bad, and the following rules do not need to be checked. In Table \ref{ex:he-rule}, we give examples for cases of each rule. In the first one, the summary is not fluent, because the patient of the predicate \zhcn{``找''} (seek for) is missing. The second summary is fluent, but the content is not related to the source, in that we cannot determine if Lei Jun is actually fighting the scalpers based on the source content. In the third one, the summary is fluent and related to the source content, but the facts are wrong, as the summary is made up by facts of different people. The last one met all the three rules, and thus it is considered good.

\begin{acks}

This work is supported in part by the \grantsponsor{GS501100001809}{National Natural Science Foundation of China}{http://dx.doi.org/10.13039/501100001809} under Grant No.~\grantnum{GS501100001809}{61673028}.

\end{acks}


\nocite{ma2018naacl,ma2018ijcai,lin2018decoding,xu2018dpgan}

\bibliographystyle{ACM-Reference-Format}
\bibliography{sample-bibliography}

\end{document}